\documentclass[conference]{IEEEtran}
\IEEEoverridecommandlockouts
\usepackage{cite}
\usepackage{amsmath,amssymb,amsfonts}
\usepackage{algorithmic}
\usepackage{algorithm}

\usepackage{graphicx}
\usepackage{textcomp}
\usepackage{xcolor}
\usepackage[table]{xcolor}
\definecolor{BestBg}{RGB}{255,224,180}
\def\BibTeX{{\rm B\kern-.05em{\sc i\kern-.025em b}\kern-.08em
    T\kern-.1667em\lower.7ex\hbox{E}\kern-.125emX}}
\begin{document}

\title{Modality Dominance-Aware Optimization for Embodied RGB–Infrared Perception}

\author{
\centering
\IEEEauthorblockN{
Xianhui Liu$^{1}$,
Siqi Jiang$^{2}$,
Yi Xie$^{3}$,
Yuqing Lin$^{1}$,
Siao Liu$^{4,5}$
}
\IEEEauthorblockA{
$^{1}$ College of Electronics and Information Engineering, Tongji University \\
$^{2}$ School of Computer Science and Technology, Tongji University \\
$^{3}$ Department of Electrical and Computer Engineering, The University of Arizona \\
$^{4}$ School of Future Science and Engineering, Soochow University \\
$^{5}$ Key Laboratory of General Artificial Intelligence and Large Models in Provincial Universities, Soochow University \\
\{xianhui\_l@163.com, 2432013@tongji.edu.cn, saliu@suda.edu.cn\}
}
}

\maketitle


\begin{abstract}
RGB–Infrared (RGB–IR) multimodal perception is fundamental to embodied multimedia systems operating in complex physical environments. Although recent cross-modal fusion methods have advanced RGB–IR detection, the optimization dynamics caused by asymmetric modality characteristics remain underexplored. 
In practice, disparities in information density and feature quality introduce persistent optimization bias, leading training to overemphasize a dominant modality and hindering effective fusion. To quantify this phenomenon, we propose the Modality Dominance Index (MDI), which measures modality dominance by jointly modeling feature entropy and gradient contribution. Based on MDI, we develop a Modality Dominance-Aware Cross-modal Learning (MDACL) framework that regulates cross-modal optimization. MDACL incorporates Hierarchical Cross-modal Guidance (HCG) to enhance feature alignment and Adversarial Equilibrium Regularization (AER) to balance optimization dynamics during fusion. Extensive experiments on three RGB–IR benchmarks demonstrate that MDACL effectively mitigates optimization bias and achieves SOTA performance.
\end{abstract}

\begin{IEEEkeywords}
Multimodal Perception, RGB-Infrared, Modality Imbalance
\end{IEEEkeywords}

\section{Introduction}
\label{sec:intro}
Multimodal perception that integrates visible (RGB) and infrared (IR) inputs is critical for embodied intelligent systems operating in complex physical environments. For real-world agents such as autonomous robots, robust object detection must be maintained under adverse conditions including low illumination and haze, where RGB perception degrades substantially. Infrared imaging complements RGB by providing stable thermal cues, making RGB–IR fusion particularly effective for reliable perception in dynamic environments. Consequently, RGB–IR detection has become a key component of embodied multimedia systems. 
Despite its potential, training a unified detector that effectively leverages both modalities remains challenging, primarily due to the significant heterogeneity in their inherent spectral domain gap and data heterogeneity.
\begin{figure}[t]
\centerline{\includegraphics[width=1.0\linewidth]{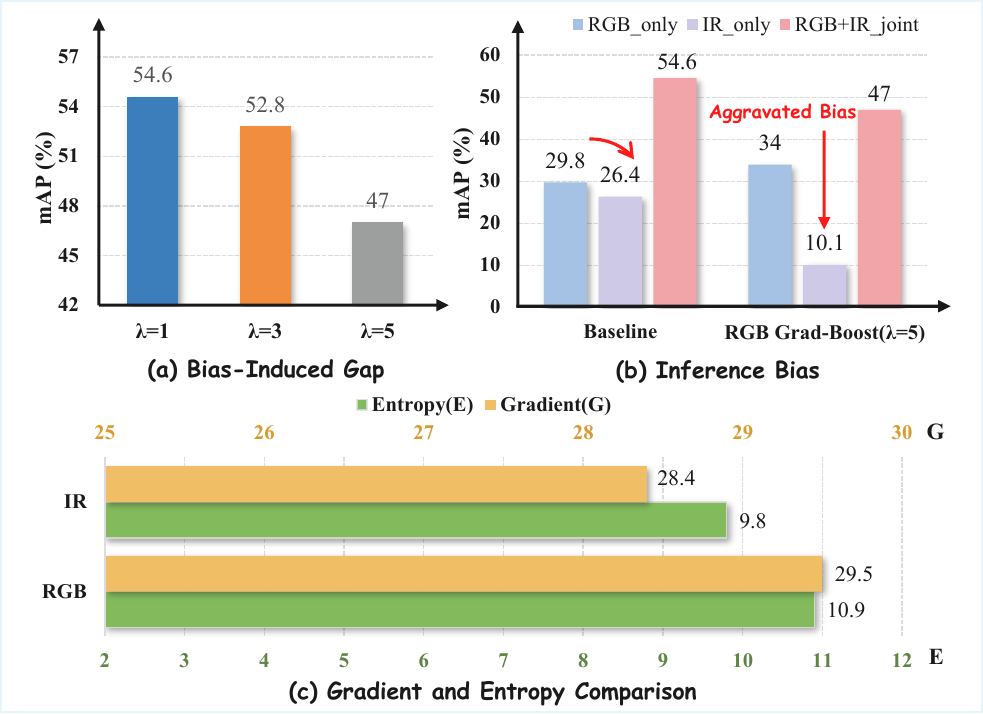}}
\vspace{-10pt}
\caption{Illustration of the optimization bias phenomenon in RGB–Infrared detection.
(a) Performance comparison on the M3FD dataset under different optimization settings, including the RGB+IR joint training baseline ($\lambda\!=\!1$) and two variants with manually amplified RGB gradients ($\lambda\!=\!3$ and $\lambda\!=\!5$). 
(b) Inference performance on M3FD using RGB-only, IR-only, and RGB+IR inputs for the baseline RGB+IR model and the RGB Grad-Boost ($\lambda\!=\!5$) variant. 
(c) Average gradient contribution and object-region feature entropy of RGB and IR modalities during joint training on M3FD.}
\label{intro}
\vspace{-20pt}
\end{figure}

A series of works have been proposed for cross-modal fusion and obtain remarkable progress, which can be broadly divided into two categorizes.
One line focuses on modeling modal consistency and complementarity\cite{cao2025cmrfusion, shen2024icafusion, wang2025rethinking, zhao2023cddfuse}. For instance, CMRFusion\cite{cao2025cmrfusion} and ICAFusion\cite{shen2024icafusion} design specialized modules to decouple shared and specific features across modalities.
Another line aims to mitigate the discrepancies between RGB and IR features in spatial or semantic representation\cite{chen2024weakly, guo2024damsdet, zhao2024equivariant}. 
Representative methods such as OAFA \cite{chen2024weakly} and DAMSDet \cite{guo2024damsdet} aim to address the mis-alignment issue by implicitly aligning RGB and IR modalities to improve detection performance.



%
Despite notable advances, most existing methods implicitly assume balanced modality contributions, leaving the optimization dynamics induced by asymmetric RGB–IR modalities largely unexplored. However, disparities in information density and feature quality often lead to persistent \textit{\textbf{optimization bias}}, causing the learning process to favor one modality.
As shown in Fig.~\ref{intro}(a), further amplifying the gradient of the dominant RGB modality on M3FD consistently degrades performance, and the degradation intensifies as the bias increases. Fig.~\ref{intro}(b) further reveals that under such biased optimization, the jointly trained detector performs markedly better with RGB-only inputs than with IR-only inputs, indicating that training has disproportionately relied on RGB modality. 
To provide a quantitative perspective, Fig.~\ref{intro}(c) shows that modality with higher entropy and stronger gradient contributions receives greater optimization preference, indicating a positive correlation between these factors and optimization bias.

Motivated by these findings, we first design a simple yet effective metric \textbf{M}odality \textbf{D}ominance \textbf{I}ndex (\textbf{MDI}) to explicitly measure optimization bias by jointly modeling feature entropy and gradient energy. 
Furthermore, we propose a novel \textbf{M}odality \textbf{D}ominance-\textbf{A}ware \textbf{C}ross-modal \textbf{L}earning (\textbf{MDACL}) framework to solve the training optimization bias problem in RGB-Infrared detection and improve its generalization performance. Specifically, the MDACL contains two key components: the Hierarchical Cross-modal Guidance (HCG) and the Adversarial Equilibrium Regularization (AER) strategy.
To enhance consistency between the two modalities, the HCG adopts a dual-stage interaction design, guiding structure-oriented alignment on low-level features and semantics-driven cross-modal consistency on high-level features, thereby effectively mitigating feature-level misalignment.
Moreover, given that prior studies\cite{wang2020makes, wei2024enhancing} have shown that optimization imbalance across modalities would lead to sub-optimal convergence behavior, we introduce AER and devise a simple yet efficient inverse weight solution to adjust the optimization dynamics, encouraging a more balanced learning process.
To validate the effectiveness of MDACL, we conduct extensive experiments on three RGB-Infrared detection benchmarks.
In summary, our contribution encompasses three main manifolds:

\begin{itemize}
\item To the best of our known, we are the first to identify and mitigate the \textbf{\textit{optimization bias}} problem in RGB-Infrared detection through comprehensive quantitative analysis.
\item We propose a novel \textbf{M}odality \textbf{D}ominance-\textbf{A}ware \textbf{C}ross-modal \textbf{L}earning (\textbf{MDACL}) framework, which consists of two core components: the Hierarchical Cross-modal Guidance (HCG), designed to enhance cross-modal consistency and mitigate misalignment; and the Adversarial Equilibrium Regularization (AER) strategy, introduced to actively regulate the optimization dynamics for a more stable cross-modal learning process.
\item Extensive experiments demonstrate that MDACL can achieve SOTA performance on three benchmarks, with considerable improvements over existing methods.
\end{itemize}

\begin{figure*}[htbp]
\centerline{\includegraphics[width=1.0\linewidth]{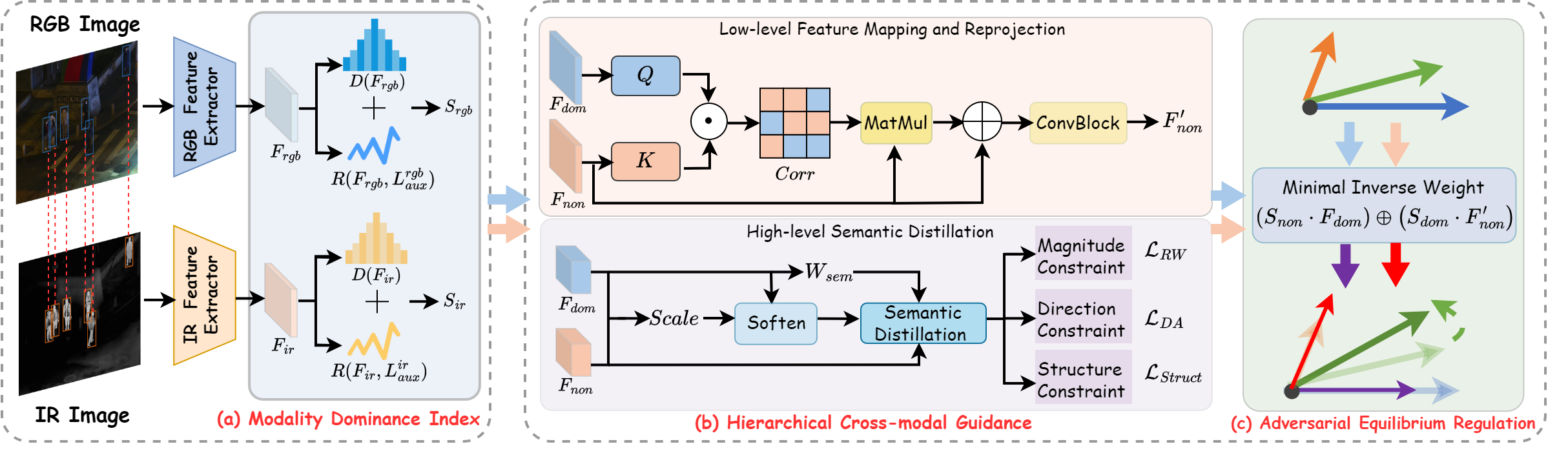}}
\vspace{-10pt}
\caption{Overview of the proposed \textbf{MDACL} framework. RGB and IR images are processed by a dual-stream backbone, followed by \textbf{(a) Modality Dominance Index (MDI)} to estimate modality dominance.
The dominance scores guide \textbf{(b) Hierarchical Cross-modal Guidance (HCG)} for cross-modal feature alignment, and \textbf{(c) Adversarial Equilibrium Regularization (AER)} for balanced feature fusion and stable optimization.}
\label{framework}
\vspace{-10pt}
\end{figure*}

\section{RELATED WORK}
\noindent \textbf{Cross-Modal Fusion for RGB-Infrared Detection.}
As a key technique in RGB–Infrared detection, cross-modal fusion has become an essential research direction.
Existing methods primarily address two challenges: (1) exploiting modality complementarity while preserving modality-specific characteristics, and (2) alleviating cross-modal misalignment.
Accordingly, prior work can be broadly categorized into two lines. The first line focuses on disentangling shared and modality-specific representations. CMRFusion~\cite{cao2025cmrfusion} explicitly models common and unique branches for each modality, while CDDFuse~\cite{zhao2023cddfuse} enforces feature decomposition through a correlation-driven loss. In addition, attention-based mechanisms are widely adopted to capture global cross-modal interactions and complementary information~\cite{shen2024icafusion, cao2023multimodal}.
The second line targets cross-modal feature alignment to mitigate spatial or semantic discrepancies. OAFA~\cite{chen2024weakly} explicitly models cross-modal spatial offsets for alignment, whereas DAMSDet~\cite{guo2024damsdet} employs deformable cross-attention to accommodate modality misalignment in complex scenes.
Despite notable progress, most existing RGB–IR fusion strategies overlook the asymmetric optimization dynamics during training.

\noindent \textbf{Optimization Dynamics in Cross-Modal Learning.}
In cross-modal learning, a key challenge is the asynchronous convergence of different modalities, which can lead to suboptimal performance\cite{wang2020makes} and unstable training process. A series of studies\cite{wei2024enhancing,zhang2024multimodal,xu2023mmcosine} argue that a better-performing modality would frequently dominate gradient updates and inadvertently suppressing the learning of the weaker one. To mitigate such gradient conflicts, recent methods introduce some regularization strategies like Pareto integration\cite{wei2024mmpareto} and adaptive gradient modulation\cite{li2023boosting}. However, the role of optimization dynamics in addressing modality imbalance remains underexplored in the domain of RGB-Infrared object detection.

\section{Problem Formulation}
Given a paired RGB-IR image $\mathbf{I} = \{\mathbf{I}^\text{rgb}, \mathbf{I}^\text{ir}\}$, the goal of RGB-IR object detection is to predict the location of target objects $\mathcal{O} = \{(\mathbf{b}_i, c_i)\}_{i=1}^N$, where $\mathbf{b}_i \in \mathbb{R}^4$ is a bounding box and $c_i \in \{1, \dots, K\}$ its class label. We aim to learn a model $\mathcal{F}_\theta$ parameterized by $\theta$ from a dataset $\mathcal{D} = \{(\mathbf{I}_j, \mathcal{O}_j)\}$. Formally, the standard objective can be formulated as follow:
\begin{equation}
    \theta^* = \arg\min_{\theta} \frac{1}{|\mathcal{D}|} \sum_{(\mathbf{I}, \mathcal{O}) \in \mathcal{D}} \mathcal{L}_\text{det}\left(\mathcal{F}_\Theta(\mathbf{I}^\text{rgb}, \mathbf{I}^\text{ir}), \mathcal{O}\right).
\end{equation}
For most RGB-IR detection frameworks, $\mathcal{F}_\theta$ contains three components: modality-specific encoders $\mathcal{G}_*$, a cross-modal fusion module $\mathcal{H}$ and a standard detection head $\mathcal{D}et$:
\begin{equation}
\mathcal{F}_\theta(I^{RGB}, I^{IR}) = \mathcal{D}et(\mathcal{H}\phi \big( \mathcal{G}\psi(I^{RGB}), \mathcal{G}\chi(I^{IR}) \big)),
\end{equation}
where $\mathcal{G}\psi$ and $\mathcal{G}\chi$ can extract RGB and Infrared features respectively. To fully utilize multimodal information, existing approaches~\cite{yuan2025unirgb, shen2024icafusion, wang2025rethinking} most focus on the design of the fusion module $\mathcal{H}_\phi$, while overlooking the inherent optimization dynamics problem. This limitation motivates us to explicitly model and balance the modality-specific learning signal.



\section{METHOD}
\subsection{Overview}
In this work, we explore RGB–Infrared detection from an optimization perspective, aiming to mitigate training bias induced by asymmetric modality characteristics. Rather than assuming balanced modality contributions, we explicitly model modality dominance and regulate cross-modal optimization accordingly. As shown in Fig.~\ref{framework}, we propose a unified dominance-aware framework that estimates modality dominance, guides cross-modal alignment, and stabilizes feature fusion under severe modality discrepancies. 
In the following, we will introduce the Modality Dominance Index in subsec.~\ref{MDI}, present the Hierarchical Cross-modal Guidance in subsec.~\ref{HCG}, and describe the Adversarial Equilibrium Regularization in subsec.~\ref{AER}.

\subsection{Modality Dominance Index}
\label{MDI}





In RGB–IR detection tasks, the inherent discrepancies across modalities often induce imbalanced optimization dynamics, ultimately hindering effective cross-modal feature cooperation. Motivated by the insight, we introduce Modality Dominance Index (MDI) to dynamically quantify the contribution of each modality during training. Let $F_i$ denote the feature map of modality $i\in \{RGB,IR\}$. The proposed MDI captures two complementary aspects of modality quality:
\noindent\textbf{(1) Representational Diversity.} To assess the inherent information richness of each modality, we define a diversity function $D(F_i)$ that measures the statistical dispersion of its characteristic activations. Modalities with richer and more uniformly distributed activations receive higher diversity scores, indicating a stronger potential contribution.

\noindent\textbf{(2) Task-Response Sensitivity.} Representational richness is insufficient to reflect modality importance. 
We therefore define a task-response function $R(F_i, L_{aux}^i)$, which evaluates how sensitive the detection task is to each modality. 
A higher response score indicates that slight perturbations induce larger changes in the detection loss, implying higher task relevance.

The Modality Dominance Index $S$ is obtained by normalizing and linearly combining the diversity and response terms: 
\begin{equation}
    S_i = \delta \cdot D(F_i)+(1-\delta) \cdot R(F_i, L_{aux}^i),
\end{equation}
where $\delta$ balances representational diversity and task-response sensitivity. The MDI computation procedure is provided in Algorithm 1. A higher MDI value would indicate that the modality is more dominant in the current training context.

\begin{algorithm}[t]
\caption{Modality Dominance Index (MDI)}
\label{alg:mdi}
\begin{algorithmic}[1]
\REQUIRE Modality features $\{F_i\}_{i\in\{rgb,ir\}}$, auxiliary detector $g(\cdot)$, ground truth $M_{GT}$, balance factor $\delta$
\ENSURE Modality dominance scores $S_{rgb}$, $S_{ir}$
\FOR{$i \in \{rgb, ir\}$}
    \STATE $D(F_i) \leftarrow \text{Entropy}(\text{Softmax}(\text{Flatten}(F_i)))$
    \STATE $L_{aux}^i \leftarrow \| g(F_i) - M_{GT} \|^2$
    \STATE $R(F_i,L_{aux}^i) \leftarrow \left\| \partial L_{aux}^i / \partial F_i \right\|_2$
\ENDFOR
\STATE Normalize $\{D(F_i)\}$ and $\{R(F_i,L_{aux}^i)\}$
\STATE $S_i \leftarrow \delta \cdot D(F_i) + (1-\delta) \cdot R(F_i,L_{aux}^i)$
\RETURN $\{S_{rgb}, S_{ir}\}$
\end{algorithmic}
\end{algorithm}

\subsection{Hierarchical Cross-modal Guidance}
\label{HCG}
\subsubsection{Low-Level Feature Mapping and Reprojection}
Owing to the spectral discrepancies between RGB and IR imaging mechanisms, low-level features which encode texture patterns and spatial structures often exhibit cross-modal misalignment.
To enhance structural-level consistency, we dynamically project the non-dominant modality feature $F_{non}$ into the structural space defined by the dominant modality $F_{dom}$, where modality dominance is determined by the MDI.

Specifically, we transform $F_{non}$ and $F_{dom}$ into low-dimensional Query ($Q$) and Key ($K$) representations using separate convolutions.
The cross-modal spatial correlation matrix $Corr \in \mathbb{R}^{HW \times HW}$ is then computed via a scaled dot-product between $Q$ and $K$, followed by Softmax normalization.
With the guidance of the correlation matrix $Corr$, we can reproject $F_{non}$ onto the structural manifold defined by $F_{dom}$, yielding an aligned representation $F_{reproj}$:
\begin{equation}
F_{reproj} = \text{MatMul}(Corr, F_{non}).
\end{equation}
To preserve modality-specific features while aligning structures, we fuse the reprojected feature $F_{reproj}$ with the original non-dominant feature $F_{non}$ via a residual addition and refine the result with a lightweight convolutional block.
\begin{equation}
F'_{non} = \text{ConvBlock}(F_{non} + F_{reproj}).
\end{equation}

\subsubsection{High-Level Semantic Distillation}
In contrast to the structure-focused low-level stage, the high-level stage aims to guide $F_{non}$ toward the semantic richness of $F_{dom}$.
To this end, we devise a cross-modal semantic distillation loss $\mathcal{L}_{Distill}$, where $F_{dom}$ serves as the teacher $F_T$ and $F_{non}$ acts as the student $F_S$.
The proposed $\mathcal{L}_{Distill}$ is constructed as a multi-objective compound loss function, targeting both semantic alignment and structural robustness preservation.


To mitigate potential harmful knowledge transfer when the modalities differ significantly, we first compute the initial feature variance $\Delta$ between $F_T$ and $F_S$ to dynamically generates a scaling factor $\text{Scale} \propto e^{-\Delta}$, which softens the teacher signal.
\begin{equation}
\Delta = \frac{1}{C H W} \parallel F_T - F_S \parallel_2^2,
\end{equation}
where $C, H$, and $W$ are the channel count, height, and width of the feature, respectively.


\noindent\textbf{Semantic Alignment Supervision.} 
To ensure that $F_S$ accurately approximates $F_T$ in the semantic space, we introduce two complementary loss terms, $\mathcal{L}_{RW}$ and $\mathcal{L}_{DA}$, which jointly constrain the consistency of feature magnitude and direction.

The Region-Weighted $L_2$ Loss $\mathcal{L}_{RW}$ constrains the magnitude of the feature by minimizing the squared difference between the channel-normalized representations. 
We further generate a weighting map $W_{sem}$ based on the activation intensity of $F_T$, assigning greater supervision to discriminative regions and allowing task-aware semantic alignment:
\begin{equation}
\mathcal{L}_{RW} = \left\| W_{sem} \odot (F_S - F_T) \right\|_2^2.
\end{equation}
The $W_{sem}$ is obtained by normalizing the channel-wise $L_2$ norm of the teacher feature $\left\| F_T \right\|_{c,2}$ with its spatial mean $\mathbb{E}(\cdot)$:
\begin{equation}
W_{sem} = \frac{\left\| F_T \right\|_{c, 2}}{\mathbb{E}\left( \left\| F_T \right\|_{c, 2} \right)}.
\end{equation}

To enforce directional consistency in the semantic dimension, the Cosine Similarity Loss $\mathcal{L}_{DA}$ minimizes the angle between the feature vectors of $F_S$ and $F_T$, ensuring that the student learns the semantic manifold of the teacher:
\begin{equation}
\mathcal{L}_{DA} = 1 - \frac{1}{N} \sum_{i} \frac{{F_S}_i \cdot {F_T}_i}{|| {F_S}_i ||_2 || {F_T}_i ||_2},
\end{equation}
where $N$ is the total number of spatial locations.

\noindent\textbf{Structural Preservation Constraint.} 
To avoid over-smoothing $F_S$ during semantic distillation, we introduce a gradient-based structural preservation term to align the spatial variation rates between $F_S$ and $F_T$, thus implicitly maintaining the structural consistency of the feature maps:
\begin{equation}
\mathcal{L}_{Struct} = \left| \text{Grad}(F_S) - \text{Grad}(F_T) \right|,
\end{equation}
where $Grad(F)$ represents the average absolute spatial gradient of feature $F$, calculated via finite difference approximation:
\begin{equation}
Grad(F) = \mathbb{E}\left[\left| F_{x} - F_{x-1} \right|\right] + \mathbb{E}\left[\left| F_{y} - F_{y-1} \right|\right].
\end{equation}
The semantic distillation loss $\mathcal{L}_{Distill}$ is the weighted summation of the aforementioned components:
\begin{equation}
\mathcal{L}_{Distill} = \alpha \mathcal{L}_{RW} + \beta \mathcal{L}_{DA} + \gamma \mathcal{L}_{Struct},
\end{equation}
where $\alpha, \beta, \gamma$ are hyperparameters that weight the constraints.



\subsection{Adversarial Equilibrium Regulation}
\label{AER}
In RGB-Infrared fusion, conventional approaches often assign larger fusion weights to the dominant modality. However, such “advantage amplification” skews the optimization dynamics, leading the network to over-rely on a single modality while degrading the contributions of the other. To mitigate such imbalance, we draw inspiration from game theory and propose the Adversarial Equilibrium Regulation (AER) strategy.

From a game-theoretic viewpoint, the two modalities can be interpreted as cooperative–competitive agents. An overly dominant modality drives the system away from an optimal joint solution, whereas maintaining a mutually regulated and complementary interaction enables the model to approach a Pareto-optimal state. The insight highlights a key principle for multimodal fusion: we should appropriately suppress the dominant modality and encourage the weaker modality to achieve a more balanced and efficient cooperative equilibrium.

Building on this insight, we design a simple yet effective instantiation — the Minimal Inverse Weight (MIW) scheme. During feature fusion, MIW leverages the Modality Dominance Index $S$ as the regulating signal and assigns higher fusion weights to the non-dominant modality $F_{non}$, while reducing the contribution of the dominant one $F_{dom}$:
\begin{equation}
F_{{fused}} = \left( S_{{non}} \cdot F_{{dom}} \right) \oplus \left( S_{{dom}} \cdot F'_{{non}} \right).
\end{equation}
The minimal inverse weighting formulation enables the network to maintain an “adversarial equilibrium” throughout backpropagation with negligible computational overhead.


Although MIW represents a straightforward instantiation of the proposed AER strategy, it effectively validates the core idea. In future work, we will investigate more advanced regulation paradigms to further strengthen the dynamic optimization equilibrium in RGB-Infrared multimodal learning.

\section{EXPERIMENTS AND ANALYSIS}

\subsection{Experimental Setup}

\noindent\textbf{Datasets and Evaluation Metrics.} We conduct experiments on three widely used RGB-Infrared detection benchmarks: LLVIP\cite{jia2021llvip}, M3FD\cite{liu2022target}, and FLIR\cite{zhang2020multispectral}. To assess overall detection performance, we adopt two commonly used metrics: mAP and mAP50, where mAP denotes the mean Average Precision (AP) averaged over IoU thresholds from 0.50 to 0.95, and mAP50 corresponds to AP at the 0.50 IoU threshold.

\noindent\textbf{Implement Details.} All experiments are conducted on NVIDIA RTX 4090 GPUs. We defaultly use the SGD optimizer with a momentum of 0.937 and a weight decay of 0.0005. The initial learning rate is set to 0.01 and gradually decayed using a cosine annealing scheduler. For data pre-processing, all input images are resized to 640×640. For a fair comparison, we report all results over five times.
\begin{table}[b]
\vspace{-15pt}
\begin{center}
\caption{COMPARISON WITH OTHER METHODS ON M3FD: BEST IN BOLD, SECOND UNDERLINED.}\label{M3FD}
\vspace{-10pt}
\resizebox{\linewidth}{!}{
\begin{tabular}{lllll}
\hline
{Model}     &   {Data Type}   &  {Backbone}    &    {mAP50 $\uparrow$} &    {mAP $\uparrow$}        \\ \hline
TarDAL\cite{liu2022target}  & IR+RGB    & CSPDarknet53 &  80.7  & 54.1    \\
CDDFuse\cite{zhao2023cddfuse}        & IR+RGB    & CSPDarknet53  & 81.2  & 53.6    \\
KCDNet\cite{wang2024kcdnet}         & IR+RGB    & CSPDarknet53  &  83.2  & 56.3     \\
DAMSDet\cite{guo2024damsdet} & IR+RGB    & ResNet50    & 80.2  & 52.9       \\
EMMA\cite{zhao2024equivariant}    & IR+RGB    & CSPDarknet53  & 82.9  & 55.4     \\
CRSIOD\cite{wang2024cross}  & IR+RGB    & CSPDarknet53   & \underline{84.0}  & \underline{57.2}     \\
YOLOv8l-IR\cite{varghese2024yolov8}     & IR        & CSPDarknet53  & 79.5  & 53.1      \\
YOLOv8l-RGB\cite{varghese2024yolov8}    & RGB       & CSPDarknet53 &  80.9  & 52.5     \\
Ours           & IR+RGB    & CSPDarknet53  & \cellcolor{BestBg}\textbf{86.8}  & \cellcolor{BestBg}\textbf{60.5}     \\ \hline
\end{tabular}
}
\end{center}
\vspace{-10pt}
\end{table}

\subsection{Experiment Results}
\noindent\textbf{Results on M3FD.}
M3FD is a widely used RGB–Infrared benchmark covering diverse scenes and severe weather conditions.
As reported in Table~\ref{M3FD}, our approach consistently outperforms all competing methods, delivering 86.8\% mAP50 and 60.5\% mAP, with clear margins of 2.8\% and 3.3\% over the previous state-of-the-art CRSIOD.
These results highlight the importance of explicitly mitigating cross-modal structural misalignment and optimization imbalance when dealing with complex environmental variations, and confirm the effectiveness of our design under the challenging conditions of M3FD.

\begin{table}[t]
\begin{center}
\caption{COMPARISON WITH OTHER METHODS ON LLVIP: BEST IN BOLD, SECOND UNDERLINED.}\label{LLVIP}
\vspace{-12pt}
\resizebox{\linewidth}{!}{
\begin{tabular}{lllll}
\hline
Model            & Data Type & Backbone     & mAP50 $\uparrow$ &  mAP $\uparrow$  \\ \hline
ICAFusion\cite{shen2024icafusion} & IR+RGB    & CSPDarknet53 & 95.2  & 60.1 \\
LUT-Fuse\cite{yi2025lut}        & IR+RGB    & CSPDarknet53 & 94.1  & 61.4 \\
Fusion-Mamba\cite{dong2025fusion}     & IR+RGB    & CSPDarknet53     & 97.0  & 64.3 \\
UniRGB-IR\cite{yuan2025unirgb}        & IR+RGB    & Transformer  & 96.1  & 63.2 \\
CSAA\cite{cao2023multimodal}    & IR+RGB    & ResNet50     & 94.3  & 54.2 \\
Text-IF\cite{yi2024text}          & IR+RGB    & Transformer  & 94.1  & 60.2 \\
FFM\cite{wang2025rethinking}   & IR+RGB    & CSPDarknet53 & \underline{97.6}  & \underline{64.8} \\
YOLOv8l-IR\cite{varghese2024yolov8}       & IR        & CSPDarknet53 & 94.6  & 61.7 \\
YOLOv8l-RGB\cite{varghese2024yolov8}      & RGB       & CSPDarknet53 & 91.8  & 53.6 \\
Ours    & IR+RGB   & CSPDarknet53 & \cellcolor{BestBg}\textbf{97.9}  & \cellcolor{BestBg}\textbf{66.5} \\ \hline
\end{tabular}
}
\end{center}
\vspace{-20pt}
\end{table}

\begin{table}[t]
\begin{center}
\caption{COMPARISON WITH OTHER METHODS ON FLIR: BEST IN BOLD, SECOND UNDERLINED.}\label{FLIR}
\vspace{-12pt}
\resizebox{\linewidth}{!}{
\begin{tabular}{lllll}
\hline
Model       & Data Type & Backbone     & mAP50 $\uparrow$ & mAP $\uparrow$  \\ \hline
ICAFusion\cite{shen2024icafusion}   & IR+RGB    & CSPDarknet53 & 79.2  & 41.4 \\
CSAA\cite{cao2023multimodal}        & IR+RGB    & ResNet50     & 79.2  & 41.3 \\
CrossFormer\cite{lee2024crossformer} & IR+RGB    & CSPDarknet53 & 79.3  & 42.1 \\
UniRGB-IR\cite{yuan2025unirgb}   & IR+RGB   & Transformer  & \underline{81.4}  & \underline{44.1} \\
FFM\cite{wang2025rethinking}         & IR+RGB    & CSPDarknet53 & \underline{81.4}  & 42.3 \\
YOLOv8l-IR\cite{varghese2024yolov8}  & IR        & CSPDarknet53 & 72.9  & 38.3 \\
YOLOv8l-RGB\cite{varghese2024yolov8} & RGB       & CSPDarknet53 & 66.3  & 28.2 \\
Ours  & IR+RGB    & CSPDarknet53 & \cellcolor{BestBg}\textbf{83.2}  & \cellcolor{BestBg}\textbf{44.6} \\
\hline
\end{tabular}
}
\end{center}
\vspace{-25pt}
\end{table}

\noindent\textbf{Results on LLVIP.}
LLVIP is a large-scale dataset collected under low-light conditions, where modality imbalance commonly arises.
As summarized in Table~\ref{LLVIP}, our method establishes new state-of-the-art results on LLVIP, achieving 97.9\% mAP50 and 66.5\% mAP, and consistently surpassing strong recent competitors such as Fusion-Mamba and FFM.
This consistent improvement can be attributed to our modality-dominance-aware learning strategy, which explicitly regulates cross-modal optimization dynamics and facilitates stable and effective feature fusion under severe modality discrepancies.

\noindent\textbf{Results on FLIR.}
FLIR is a real-world RGB–Infrared benchmark characterized by cluttered backgrounds and pronounced appearance gaps between modalities.
As shown in Table~\ref{FLIR}, our approach delivers superior detection accuracy on FLIR, reaching 83.2\% mAP50 and 44.6\% mAP, outperforming all existing methods.
In contrast to ICAFusion, which performs feature interaction under the implicit assumption of balanced modality contributions, our approach explicitly identifies and regulates the asymmetric optimization behavior between RGB and IR modalities, thereby enabling more effective feature alignment and leading to improved detection performance.

\subsection{Qualitative Analysis}
\begin{figure}[t]
\centering
\centerline{\includegraphics[width=1.05\linewidth]{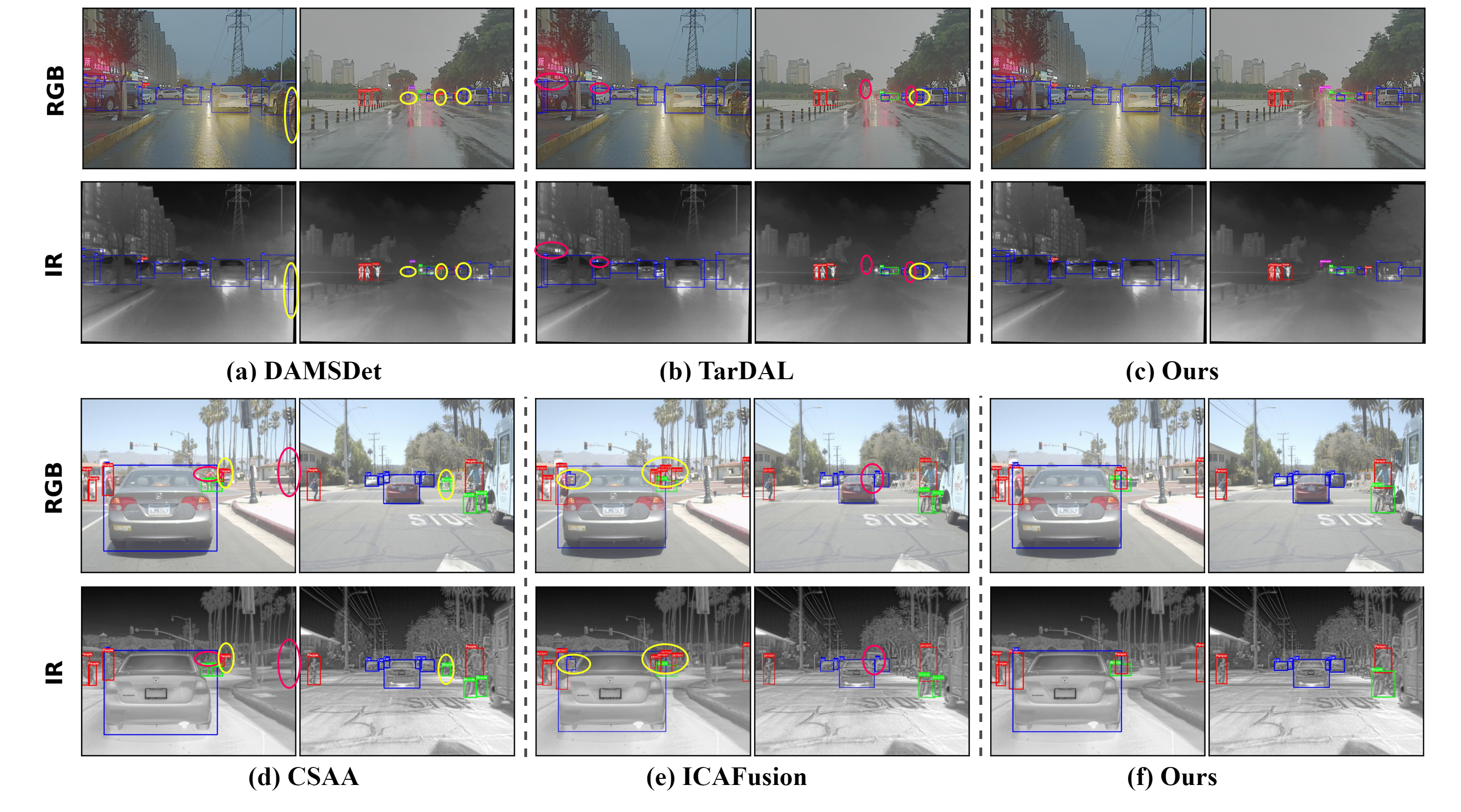}}
\vspace{-10pt}
\caption{Visualization of some RGB-Infrared detection methods on M3FD and FLIR. (a)-(c)
present the results of M3FD dataset, and (d)-(f) present the results of FLIR dataset. The targets encircled by yellow ellipses are false positives, while those encircled by red ellipses are missed detections.}
\label{result}

\end{figure}

\noindent{\textbf{Sample Visualization.}} We visualize qualitative detection results and compare them with representative methods~\cite{guo2024damsdet, liu2022target, shen2024icafusion, cao2023multimodal}. As illustrated in Fig.~\ref{result}, our approach consistently reduces missed detections and false positives under challenging scenarios, including low illumination and severe occlusion, demonstrating robust and accurate detection performance.

\begin{figure}[t]
\vspace{-12pt}
\centerline{\includegraphics[width=1.0\linewidth]{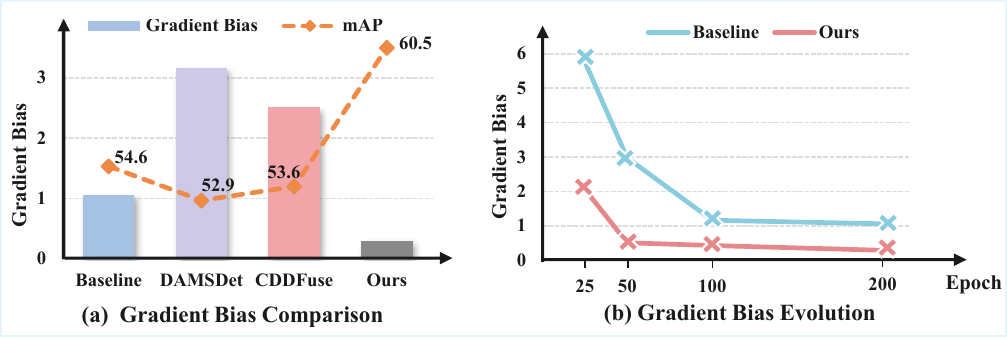}}
\vspace{-10pt}
\caption{Gradient bias comparison and its impact on model performance.
(a) Performance versus average gradient bias for four RGB–IR detectors.
(b) Evolution of average gradient bias across different training epochs.}
\label{grad}
\vspace{-10pt}
\end{figure}

\noindent{\textbf{Gradient Comparison.}}
To evaluate the effectiveness of our method in alleviating optimization bias, we analyze the \emph{gradient bias} during training and compare it with representative baselines. The gradient bias is defined as the average absolute difference between the gradient contributions of the RGB and IR branches, computed over every 100 training steps and then averaged across the full training process.
As shown in Fig.~\ref{grad}(a), detection performance exhibits a clear inverse correlation with gradient bias. In particular, our method attains the lowest gradient bias while delivering the highest performance, indicating a more balanced optimization behavior.
Fig.~\ref{grad}(b) further shows the gradient bias evolution across training stages. Compared with the baseline, our approach consistently maintains lower gradient bias throughout training, demonstrating its robustness in regulating cross-modal optimization dynamics and facilitating effective RGB–IR fusion.


\begin{table}[h]
\vspace{-15pt}
\begin{center}
\caption{ABLATION STUDY WITH CONFIGURATIONS ON LLVIP}\label{ablation}
\vspace{-10pt}
\begin{tabular}{l|cc|l|ll}
\hline
                       & \multicolumn{2}{c|}{HCG}                                                   &     &       &      \\
MDI                    & \multicolumn{1}{l}{Low-Map} & \multicolumn{1}{l|}{High-Distill} & MIW & mAP50 $\uparrow$ & mAP $\uparrow$  \\ \hline
                       & \multicolumn{1}{l}{}              & \multicolumn{1}{l|}{}                   &     & 95.5  & 63.4 \\
\multicolumn{1}{c|}{$\checkmark$} & $\checkmark$                                 &                                         &     & 96.3 \textcolor{red}{+0.8}  & 64.6 \textcolor{red}{+1.2} \\
\multicolumn{1}{c|}{$\checkmark$} &                                   & $\checkmark$                                       &     & 96.5 \textcolor{red}{+1.0}  & 64.8 \textcolor{red}{+1.4} \\
\multicolumn{1}{c|}{$\checkmark$} & $\checkmark$                                 & $\checkmark$                                       &     & 97.0 \textcolor{red}{+1.5}  & 65.2 \textcolor{red}{+1.8} \\
\multicolumn{1}{c|}{$\checkmark$}                    &                                   &                                         & $\checkmark$   & 96.4 \textcolor{red}{+0.9}  & 64.5 \textcolor{red}{+1.1} \\ 
\multicolumn{1}{c|}{$\checkmark$}                    & $\checkmark$                                 & $\checkmark$                                       & $\checkmark$   & \cellcolor{BestBg}\textbf{97.9} \textcolor{red}{+2.4}  & \cellcolor{BestBg}\textbf{66.5} \textcolor{red}{+3.1} \\ \hline
\end{tabular}
\vspace{-20pt}
\end{center}
\end{table}

\subsection{Ablation Study}

\noindent\textbf{Core Components.}
Table \ref{ablation} summarizes the ablation results on LLVIP. Since MDI quantifies modality dominance, it is designed to operate jointly with HCG or MIW.
The two HCG guidance mechanisms respectively enhance structural- and semantic-level alignment, each yielding an improvement of approximately +1\% mAP50 when applied in isolation, and delivering additional performance gains when jointly enabled.
Meanwhile, MIW rebalances the modality contributions during optimization, offering additional consistent improvements. Integrating MDI, HCG, and MIW leads to the best overall performance, substantially outperforming the baseline and highlighting the effectiveness and synergy of all components.

\begin{figure}[t]
\vspace{-15pt}
\centerline{\includegraphics[width=1.0\linewidth]{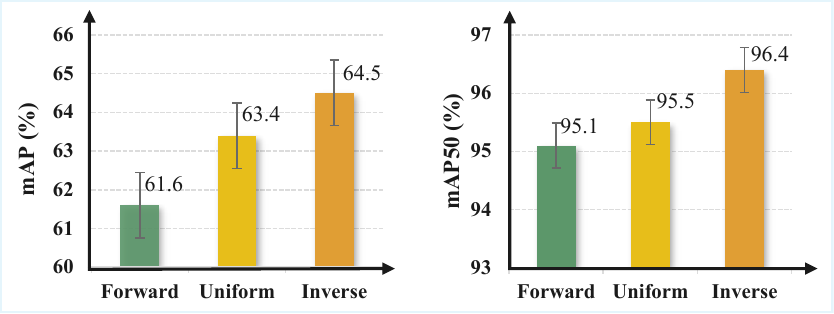}}
\vspace{-15pt}
\caption{Comparison of three modality weighting strategies on LLVIP: Forward (forward weighting), Uniform (uniform weighting used as the baseline), and Inverse (inverse weighting proposed in our work).}
\label{IWA}
\vspace{-20pt}
\end{figure}

\noindent\textbf{Weight Allocation Strategies.}
The comparison of the three weight allocation strategies is shown in Fig. \ref{IWA}. The baseline model with uniform weighting achieves 95.5\% mAP50 and 63.4\% mAP. Adopting the forward allocation strategy, which further emphasizes the dominant modality, leads to a performance drop to 95.1\% mAP50 and 61.6\% mAP, indicating that favoring the dominant modality can hinder effective learning. In contrast, the inverse allocation strategy significantly boosts performance to 96.4\% mAP50 and 64.5\% mAP. These results confirm that rebalancing optimization by strengthening the weaker modality effectively alleviates training-induced modality imbalance and yields more robust detection performance.

\section{Conclusion}
In this work, we revisit RGB–Infrared detection from an optimization-centric perspective, highlighting how asymmetric modality characteristics influence multimodal training dynamics. Empirical results show that dominant modalities tend to attract disproportionate optimization focus, hindering effective cross-modal fusion.
To quantify this behavior, we introduce the Modality Dominance Index, a concise and interpretable measure of modality-level optimization imbalance. Building on this insight, we propose the MDACL framework, which combines hierarchical cross-modal guidance with equilibrium-aware regularization to jointly align representations and balance optimization. Extensive experiments across multiple benchmarks demonstrate the effectiveness and robustness of our approach.
We believe this work underscores the importance of optimization-aware modeling for embodied multimodal perception beyond RGB–IR detection.

\bibliographystyle{IEEEbib}
\bibliography{paper}

\end{document}